\documentclass{article}

\usepackage{microtype}
\usepackage{graphicx}
\usepackage{subfigure}
\usepackage{booktabs} 

\usepackage{hyperref}
\usepackage{xcolor}
\usepackage{algorithmic}
\usepackage{algorithm}
\usepackage[algo2e,linesnumbered]{algorithm2e}
\usepackage{graphicx}
\usepackage{multirow}
\usepackage{amssymb}
\usepackage{epstopdf}
\usepackage{multicol}
\usepackage{tabularx}
\usepackage{arydshln}
\usepackage{multirow}
\usepackage{caption}
\usepackage{subfigure}
\usepackage{enumitem}
\usepackage{enumitem}
\usepackage{booktabs}
\usepackage{bm}
\usepackage{wrapfig}
\usepackage{amsmath}
\usepackage{pifont}
\usepackage{amssymb}
\usepackage{lipsum}
\usepackage{pifont}
\usepackage{amsfonts}
\usepackage{dsfont}
\usepackage{mathrsfs}
\usepackage{amsthm}
\usepackage{float}
\usepackage{multirow}
\usepackage{wrapfig}

\usepackage[accepted]{icml2022}

\usepackage{amsmath}
\usepackage{amssymb}
\usepackage{mathtools}
\usepackage{amsthm}

\usepackage[capitalize,noabbrev]{cleveref}

\theoremstyle{plain}

\theoremstyle{definition}

\theoremstyle{remark}

\usepackage{amsmath,amsfonts,bm}

\def\eqref#1{equation~\ref{#1}}

\def\1{\bm{1}}

\DeclareMathAlphabet{\mathsfit}{\encodingdefault}{\sfdefault}{m}{sl}
\SetMathAlphabet{\mathsfit}{bold}{\encodingdefault}{\sfdefault}{bx}{n}

\usepackage[textsize=tiny]{todonotes}

\icmltitlerunning{Estimating Instance-dependent Bayes-label Transition Matrix using a Deep Neural Network}

\begin{document}

\twocolumn[
\icmltitle{Estimating Instance-dependent Bayes-label Transition Matrix \\ using a Deep Neural Network}




\begin{icmlauthorlist}
\icmlauthor{Shuo Yang}{UTS}
\icmlauthor{Erkun Yang}{xd}
\icmlauthor{Bo Han}{hkbu}
\icmlauthor{Yang Liu}{ucsz}
\icmlauthor{Min Xu}{UTS}
\icmlauthor{Gang Niu}{riken}
\icmlauthor{Tongliang Liu}{usyd}

\end{icmlauthorlist}

\icmlaffiliation{UTS}{University of Technology Sydney}
\icmlaffiliation{xd}{Xidian University}
\icmlaffiliation{usyd}{TML Lab, Sydney AI Centre, The University of Sydney}
\icmlaffiliation{hkbu}{Hong Kong Baptist University}
\icmlaffiliation{ucsz}{Computer Science and Engineering, UC Santa Cruz}
\icmlaffiliation{riken}{RIKEN Center for Advanced Intelligence Project}

\icmlcorrespondingauthor{Tongliang Liu}{tongliang.liu@sydney.edu.au}

\icmlkeywords{Machine Learning, ICML}

\vskip 0.3in
]

\printAffiliationsAndNotice{} 

\begin{abstract}

In label-noise learning, estimating the \textit{transition matrix} is a hot topic as the matrix plays an important role in building \textit{statistically consistent classifiers}. Traditionally, the transition from clean labels to noisy labels (\textit{i.e.}, \textit{clean-label transition matrix (CLTM)}) has been widely exploited to learn a \textit{clean label classifier} by employing the noisy data. Motivated by that classifiers mostly output \textit{Bayes optimal labels} for prediction, in this paper, we study to directly model the transition from \textit{Bayes optimal labels} to noisy labels (\textit{i.e.}, \textit{Bayes-label transition matrix (BLTM)}) and learn a classifier to predict \textit{Bayes optimal labels}. Note that given only noisy data, it is \textit{ill-posed} to estimate either the \textit{CLTM} or the \textit{BLTM}. 
But favorably, Bayes optimal labels have less uncertainty compared with the clean labels, \textit{i.e.}, the \textit{class posteriors} of Bayes optimal labels are \textit{one-hot vectors} while those of clean labels are not. This enables two advantages to estimate the \textit{BLTM}, \textit{i.e.}, (a) a set of examples with theoretically guaranteed Bayes optimal labels can be collected out of noisy data; (b) the feasible solution space is much smaller. By exploiting the advantages, we estimate the BLTM parametrically by employing a \textit{deep neural network}, leading to better generalization and superior classification performance.
\end{abstract}

\section{Introduction}
The study of classification in the presence of noisy labels has been of interest for three decades \citep{angluin1988learning}, but becomes more and more important in weakly supervised learning \citep{thekumparampil2018robustness,li2019gradient,guo2018curriculumnet,xiao2015learning,zhang2017understanding,yang_wu_liu_xu_2021,yang2021free}. The main reason behind this is that datasets are becoming bigger and bigger. To improve annotation efficiency, these large-scale datasets are often collected from crowdsourcing platforms \citep{yan2014learning}, online queries \citep{blum2003noise}, and image engines \citep{li2017learning}, which suffer from unavoidable label noise \citep{yao2020searching}. Recent researches show that the label noise significantly degenerates the performance of deep neural networks, since deep models easily memorize the noisy labels~\citep{zhang2017understanding,yao2020searching}.

Generally, the algorithms for combating noisy labels can be categorized into \textit{statistically inconsistent algorithms} and \textit{statistically consistent algorithms}. The statistically inconsistent algorithms are heuristic, such as selecting possible clean examples to train the classifier~\citep{han2020sigua,yao2020searching,yu2019does,han2018co,malach2017decoupling,ren2018learning,jiang2018mentornet}, re-weighting examples to reduce the effect of noisy labels~\citep{ren2018learning}, correcting labels~\citep{ma2018dimensionality,kremer2018robust,tanaka2018joint,wang2022reliable}, or adding regularization~\citep{han2018masking,guo2018curriculumnet,veit2017learning,vahdat2017toward,li2017learning,li2019gradient,wu2020class2simi}. These approaches empirically work well, but there is no theoretical guarantee that the learned classifiers can converge to the optimal ones learned from clean data. 
To address this limitation, algorithms in the second category aim to design \textit{classifier-consistent} algorithms~\citep{yu2017transfer,zhang2018generalized,kremer2018robust,liu2016classification,northcuttlearning,scott2015rate,natarajan2013learning,goldberger2016training,patrini2017making,thekumparampil2018robustness,yu2018learning,liu2019peer,xu2019l_dmi,xia2020parts}, where classifiers learned on noisy data will \textit{asymptotically converge} to the optimal classifiers defined on the clean domain.

\begin{figure*}[t!]
    \centering
    \vspace{3mm}
    \resizebox{!}{0.08\linewidth}{
    \includegraphics{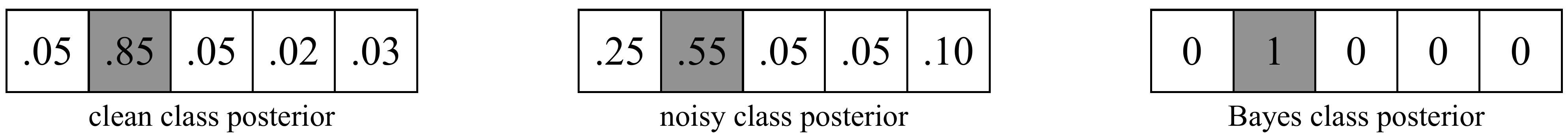}}
    \caption{The noisy class posterior is learned from noisy data. Bayes optimal label can be inferred from the noisy class posterior if the noisy rate is controlled. Also, the Bayes optimal label is less uncertain since the Bayes class posterior is \textit{one-hot} vector.}
    \label{fig:posterior}
\end{figure*}

The \textit{label transition matrix} $ T(\mathbf{x})$ plays an important role in building \textit{statistically consistent} algorithms. Traditionally, the transition matrix $ T(\mathbf{x})$ is defined to relate clean distribution and noisy distribution, where $T(\mathbf{x}) = P(\tilde{Y} \mid Y,X=\mathbf{x})$ and $X$ denotes the random variable of instances/features, $\tilde{Y}$ as the variable for the noisy label, and $Y$ as the variable for the clean label. The above matrix is denoted as the \textit{clean-label transition matrix}, which is widely used to learn a \textit{clean label classifier} by employing the noisy data. The learned clean label classifier is expected to predict a probability distribution over a set of pre-defined classes given an input, \textit{i.e.} \textit{clean class posterior probability} $P(Y\mid X)$. The clean class posterior probability is the distribution from which \textit{clean labels} are sampled. However, \textit{Bayes optimal labels} $Y^*$, \textit{i.e.}, the class labels that maximize the clean class posteriors $Y^{*} \mid {X}:= \arg \max _{Y} {P}(Y \mid {X})$, are mostly used as the predicted labels and for computing classification accuracy. 
Motivated by this, in this paper, we propose to directly model the transition matrix $T^*(\mathbf{x})$ that relates \textit{Bayes optimal distribution} and \textit{noisy distribution}, \textit{i.e.}, $T^*(\mathbf{x}) = P(\tilde{Y} \mid Y^*,X=\mathbf{x})$, where $Y^*$ denotes the variable for \textit{Bayes optimal label}. The \textit{Bayes optimal label classifier} can be learned by exploiting the Bayes-label transition matrix directly. 

Studying the transition between Bayes optimal distribution and noisy distribution is considered advantageous to that of studying the transition between clean distribution and noisy distribution. The main reason is due to that the \textit{class posteriors} of \textit{Bayes optimal labels} are \textit{one-hot vectors} while those of clean labels are not. Two advantages can be introduced by this to better estimate the instance-dependent transition matrix: \textit{\textbf{(a) A set of examples with theoretically guaranteed Bayes optimal labels can be collected out of noisy data}}. The intrinsic reason that Bayes optimal labels can be inferred from the noisy data while clean labels cannot is that the Bayes optimal labels are \textit{deterministic} while clean labels are \textit{stochastic}; the Bayes optimal labels are the labels that \textit{maximize} the \textit{clean class posteriors} while clean labels are sampled from the \textit{clean class posteriors}. 
In the presence of label noise, the labels that \textit{maximize} the \textit{noisy class posteriors} could be identical to those that \textit{maximize} the \textit{clean class posteriors} (Bayes optimal labels) under mild conditions; \textit{e.g.,} see, ~\citet{cheng2017learning}.
Therefore some instances' Bayes optimal labels can be inferred from their \textit{noisy class posteriors} while their clean labels are impossible to infer since the \textit{clean class posteriors} are unobservable, as shown in Figure~\ref{fig:posterior}. \textit{\textbf{{(b) The feasible solution space of the Bayes-label transition matrix is much smaller than that of the clean-label transition matrix.}}} This is because that Bayes optimal labels have less uncertainty compared with the clean labels. The transition matrix defined by Bayes optimal labels and the noisy labels therefore has less hypothesis complexity~\cite{liu2017algorithmic}, and can be estimated more efficiently with the same amount of training data.

These two advantages naturally motivate us to collect a set of examples and exploit their Bayes optimal labels to approximate the \textit{Bayes-label transition matrix} $T^*(\mathbf{x})$. Due to the high complexity of the instance-dependent matrix $T^*(\mathbf{x})$, we simplify its estimation by parameterizing it using a deep neural network. 
The collected examples, inferred Bayes optimal labels, and their noisy labels are served as data points to optimize the deep neural network to approximate the $T^*(\mathbf{x})$. Compared with the previous method~\citep{xia2020part}, which made assumptions and leveraged hand-crafted priors to approximate the instance-dependent transition matrices, we train a deep neural network to estimate the \textit{instance-dependent label transition matrix} with a reduced feasible solution space, which achieves lower approximation error, better generalization, and superior classification performance.

Before delving into details, we summarize our main contributions as below:

\begin{itemize}
    \item In instance-dependent label-noise learning, compared with the clean-label transition matrix, this paper proposes to study the transition probabilities between Bayes optimal labels and noisy labels, \textit{i.e.}, \textit{Bayes-label transition matrix}, which is easier to be parametrically learned because of the certainty and accessibility of the Bayes optimal labels.
    \item This paper proposes to leverage a deep neural network to capture the noisy patterns and generate the transition matrix for each input instance; it is the \textit{first} one that estimates the instance-dependent label transition matrix in a parametric way.
    
    \item The effectiveness of the proposed method is verified on three synthetic noisy datasets and a large-scale real-world noisy dataset, significant performance improvements on both synthetic and real-world noisy datasets and all experiment settings are achieved.

\end{itemize}


\section{Related Work}\label{sec:2}

\textbf{Noise model.} Currently, there are several typical label noise models. Specifically, the random classification noise (RCN) model assumes that clean labels flip randomly with a constant rate \citep{biggio2011support,manwani2013noise,natarajan2013learning}. The class-conditional label noise (CCN) model assumes that the flip rate depends on the latent clean class \citep{patrini2017making,xia2019revision,ma2018dimensionality}. The instance-dependent label noise (IDN) model considers the most general case of label noise, where the flip rate depends on its instance/features~\citep{cheng2017learning,xia2020part,zhu2020second}. Obviously, the IDN model is more realistic and applicable. For example, in real-world datasets, an instance whose feature contains less information or is of poor quality may be more prone to be labeled wrongly. The bounded instance dependent label noise (BIDN)~\citep{cheng2017learning} is a reasonable extension of IDN, where the flip rates are dependent on instances but upper bounded by a value smaller than 1. However, with only noisy data, it is a \textit{non-trivial} task to model such realistic noise without any assumption \citep{xia2020part}. This paper focuses on the challenging BIDN problem setting.

\textbf{Learning clean distributions.} It is significant to reduce the side effect of noisy labels by inferring clean distributions statistically. The label transition matrix plays an important role in such an inference process, which is used to denote the probabilities that clean labels flip into noisy labels. We first review prior efforts under the class-dependent condition \citep{patrini2017making}. By exploiting the class-dependent transition matrix $T$, the training loss on noisy data can be corrected to be an unbiased estimation of the loss on clean data. The transition matrix $T$ can be estimated in many ways, e.g., by introducing the anchor point assumption \citep{liu2016classification}, by exploiting clustering \citep{zhu2021clusterability}, by minimizing volume of $T$ \citep{li2021provably}, and by using extra clean data \citep{hendrycks2018glc,shu2020meta}. To make the estimation more accurately, a slack variable \citep{xia2019revision} or a multiplicative dual $T$ \citep{yao2020dual} can be introduced to revise the transition matrix. As for instance-dependent label-noise learning, the instance-dependent transition matrix is hard to be estimated since it relies on the unique pattern in each input instance. To learn the clean distribution from the noisy data, existing methods rely on various assumptions, e.g., the noise rate is bounded \citep{cheng2017learning}, the noise only depends on the parts of the instance \citep{xia2020part}, and additional valuable information is available \citep{berthon2020idn}. Although the above advanced methods achieve superior performance empirically, the introduction of strong assumptions limit their applications in practice. In this paper, we propose to infer Bayes optimal distribution instead of clean distribution, as Bayes optimal distribution is less uncertain and easy to be inferred under mild conditions.

\textbf{Other approaches.} Other methods exist with more sophisticated training frameworks or pipelines, including but not limited to robust loss functions \citep{zhang2018generalized,xu2019l_dmi,liu2019peer}, sample selection \citep{han2018co,wang2019co,lyu2020curriculum}, label correction \citep{tanaka2018joint,zhang2021learning,zheng2020error}, (implicit) regularization \citep{xia2021robust,zhang2017mixup,liu2020early}, semi-supervised learning \citep{nguyen2020self}, and the combination of semi-supervised learning, MixUp, regularization and Gaussian Mixture Model~\cite{li2020dividemix}. 

\section{Preliminaries}\label{sec:pre}
We introduce the problem setting and some important definitions in this section. 

\textbf{Problem setting.} This paper focuses on a classification task given a training dataset with Instance Dependent Noise (IDN), which is denoted by $\tilde{S}=\{({\mathbf{x}}_i,\tilde{y}_i)\}_{i=1}^n$. We consider that training examples $\{({\mathbf{x}}_i,\tilde{y}_i)\}_{i=1}^n$ are drawn according to random variables $({{X}},\tilde{Y})\sim\tilde{\mathcal{D}}$, where $\tilde{\mathcal{D}}$ is a noisy distribution. The noise rate for class $y$ is defined as $\rho_{y}(\mathbf{x}) = P(\tilde{Y} = y \mid Y \neq y, \mathbf{x})$. This paper focuses on a reasonable IDN setting that the noise rates have upper bounds $\rho_{max}$ as in~\citep{cheng2017learning}, \textit{i.e.}, $\forall{\mathbf{\mathbf(x)}} \in \mathcal{X}$, $0\leq \rho_{y}(\mathbf{x}) \leq \rho_{max }<1$. 
Our aim is to learn a robust classifier only from the noisy data, which could assign accurate labels for test data. 

\textbf{Clean distribution.} For the observed noisy training examples, all of them have corresponding clean labels, which are \textit{unobservable}. The clean training examples are denoted by $S=\{({\mathbf{x}}_i,y_i)\}_{i=1}^n$, which are considered to be drawn according to random variables $({{X}},Y)\sim\mathcal{D}$. The term $\mathcal{D}$ denotes the underlying clean distribution.

\textbf{Bayes optimal distribution.} Given ${X}$, its \textit{Bayes optimal label} is denoted by $Y^{*}$, $Y^{*} \mid {X}:= \arg \max _{Y} {P}(Y \mid {X}),({X}, Y) \sim \mathcal{D}$. The distribution of $(X,Y^*)$ is denoted by $\mathcal{D}^*$. Note the Bayes optimal distribution $\mathcal{D}^*$ is different from the clean distribution $\mathcal{D}$ when $P(Y|X) \notin \{0,1\}$. Like clean labels, Bayes optimal labels are unobservable due to the information encoded between features and labels is corrupted by label noise \citep{zhu2020second}. Note that it is a \textit{non-trivial task} to infer $\mathcal{D}^*$ only with the noisy training dataset $\tilde{S}$. Also, the noisy label $\tilde{y}$, clean label $y$, and
Bayes optimal label $y^*$, for the same instance ${\mathbf{x}}$ may disagree with each other \citep{cheng2017learning}.

\begin{figure*}[t!]
    \centering
    \resizebox{!}{0.265\linewidth}{
    \includegraphics{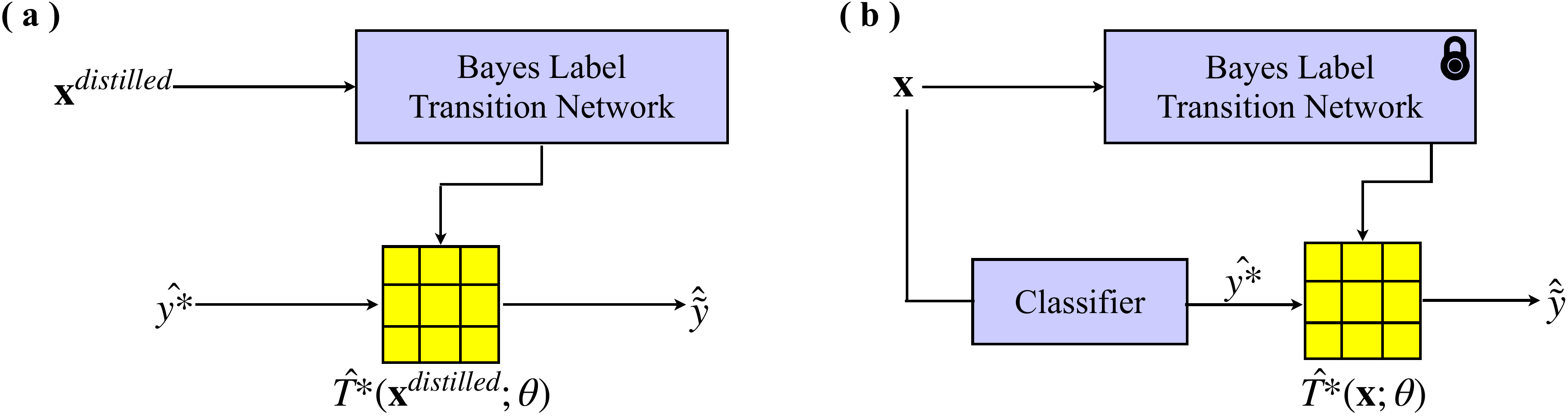}}
    \caption{(a) The Bayes Label Transition Network is used to predict \textit{Bayes-label transition matrix} for each input instance, it is trained in a supervised way by employing the collected Bayes optimal labels. (b) The learned Bayes Label Transition Network is \textit{fixed} to train the classifier by leveraging the loss correction approach~\cite{patrini2017making}. }
    \label{fig:pipeline}
    \vspace{-2mm}
\end{figure*}

\textbf{Other definitions.} The classifier is defined as $f:\mathcal{X}\rightarrow\mathcal{Y}$, where $\mathcal{X}$ and $\mathcal{Y}$ denote the instance and label spaces respectively. Let $\mathds{1}[\cdot]$ be the indicator function. Define the \textit{0-1 risk} of $f$ as $\mathds{1}(f({{X}}),Y)\triangleq\mathds{1}[f({{X}})\neq Y]$. Define the \textit{Bayes optimal classifier} $f^*$ as $f^*\triangleq\arg\min_{f}\mathbb{E}[\mathds{1}(f({{X}}),Y)]$. Note that there is NP-hardness of minimizing the 0-1 risk, which is neither
convex nor smooth \citep{bartlett2006convexity}. We can use the \textit{softmax cross entropy loss} as the \textit{surrogate loss function} to approximately learn the Bayes optimal classifier \citep{bartlett2006convexity,cheng2017learning}. We aim to learn a classifier $f$ from the noisy distribution $\tilde{\mathcal{D}}$ which also approximately minimizes $\mathbb{E}[\mathds{1}(f({{X}}),Y)]$.

\section{Method}\label{sec:method}
In \textit{instance-dependent label-noise learning}, the transition matrix is \textit{unique} and needed to be estimated for each input instance, with a carefully consideration on its ambiguous patterns. Therefore, a parametric way for transition matrix estimation benefits efficient instance-dependent noisy pattern learning. However, traditional clean-label transition matrix (\textit{clean labels $\rightarrow$ noisy labels}) is hard to be parametrically learned due to the inaccessibility of clean labels. 

To go beyond the limitation of clean-label transition matrix, this paper considers the transition between \textit{Bayes optimal labels} and \textit{noisy labels}, \textit{i.e.}, Bayes-label transition matrix (Section.~\ref{sec:bayesmatrix}). Compared with the uncertain clean labels that are hard to be collected from the noisy dataset, \textit{Bayes optimal labels} have no uncertainty and can be easily inferred from only noisy data~\cite{kremer2018robust,cheng2017learning,cheng2021learning,DBLP:conf/aaai/ZhengAD21,DBLP:conf/cvpr/WangHKCR21}. By employing some existing techniques~\cite{cheng2017learning,cheng2021learning}, a set of examples with both theoretically guaranteed Bayes optimal labels and the noisy labels can be collected out of the noisy dataset (Section.~\ref{sec:collecting}). Then, we design a parametric \textit{Bayes label transition network} to extract image patterns and estimate the instance-dependent label transition matrix (Section.~\ref{sec:estimating}), the Bayes label transition network is trained in a supervised way by employing the collected Bayes optimal labels. Finally, we combine the learned Bayes label transition network to the classifier training (Section.~\ref{sec:training}).

\subsection{Bayes-label transition matrix}\label{sec:bayesmatrix}
This paper focus on studying the transition from \textit{Bayes optimal labels} to \textit{noisy labels}. We introduce the definition of the Bayes-label transition matrix that bridges the Bayes optimal distribution and noisy distribution as follows,
\begin{equation}
    T^*_{i,j}(X) = P( \tilde{Y} = j \mid Y^* = i, X),
\end{equation}
where $T^*_{i,j}(X)$ denotes the $(i,j)$-th element of the matrix $T^*(X)$, indicating the probability of a Bayes optimal label $i$ flipped to noisy label $j$ for input $X$.

Given the \textit{noisy class posterior probability} $P(\tilde{\mathbf{Y}} \mid X=\mathbf{x})=[P(\tilde{Y}=1 \mid X=\mathbf{x}), \ldots, P(\tilde{Y}=C \mid X=\mathbf{x})]$ (which can be learned from noisy data) and the Bayes-label transition matrix $T^*_{ij}({\mathbf{x}})=P(\tilde{Y}=j|Y^*=i,X={\mathbf{x}})$, the \textit{Bayes class posterior probability} $P(\mathbf{Y^*}|X=\mathbf{x})$ can be inferred, \textit{i.e.}, $P(\mathbf{Y^*} \mid X=x)=\left(T^*(X=x)^{\top}\right)^{-1} P(\tilde{\mathbf{Y}} \mid X=x)$.

\subsection{Collecting Bayes Optimal Labels}\label{sec:collecting}
Favoured by the characteristics of deterministic, some examples' Bayes optimal labels can be inferred from the noisy class posterior probabilities automatically~\cite{cheng2017learning,cheng2021learning}. We leverage the noisy dataset distillation method in~\cite{cheng2017learning} (Theorem 2 therein) to collect a set of \textit{distilled examples} $({\mathbf{x}},\tilde{y}, \hat{y^*})$ out of the noisy dataset, where the $\tilde{y}$ is the noisy label and $\hat{y^*}$ is the inferred \textit{theoretically guaranteed} Bayes optimal label. Specifically, we can obtain distilled examples by collecting all noisy examples $(\mathbf{x}, \tilde{y})$ whose $\mathbf{x}$ satisfies $\tilde{\eta}_{{y}}(\mathbf{x}) > \frac{1+\rho_{max}}{2}$ and then assigning the label ${y}$ to it as its inferred Bayes optimal label $\hat{y^*}$, where $\tilde{\eta}_{{y}}(\mathbf{x})$ is the noisy class posterior probability of $\mathbf{x}$ on $y$ and the $\rho_{max}$ is the noise rate upper bound.
The inferred Bayes optimal label $\hat{y^*}$ may disagree with the noisy label $\tilde{y}$. The noisy class posterior probability $\tilde{\eta}$ can be estimated as $\hat{\tilde{\eta}}$ by several probabilistic classification methods (logistic regression or deep neural networks). Please refer to~\cite{cheng2017learning} for more details about the Bayes optimal label collection and the theoretical guarantee. Note that our method can also built on top of any other methods that can collect Bayes optimal labels from noisy dataset. Note also that many existing methods extracting confident examples and correcting labels~\cite{tanaka2018joint,zhang2021learning,zheng2020error} have closely relationships with Bayes optimal labels.

\begin{algorithm}[t!]
\KwIn{Noisily-labeled dataset $\tilde{S}=\{({\mathbf{x}}_i,\tilde{y}_i)\}_{i=1}^n$}
{
\textbf{Required}: the noise rate upper bound $\rho_{max}$, random initialized Bayes label transition network $\hat{T^*}(\cdot ; \theta)$, random initialized classification network $f(\cdot ; w)$}

\textcolor{blue}{// Section.~\ref{sec:collecting}: Collecting Bayes Optimal Labels}

Initialize the distilled dataset $\mathcal{S^*} = \{\}$;

Learn $\hat{\tilde{\eta}}$ on the noisy dataset $\tilde{S}$;

\For{(${\mathbf{x}}_i,\tilde{y}_i)$ in $\tilde{S}$}
{
\For{$y$ in $\mathcal{Y}$}{
    \If{$\hat{\tilde{\eta}}_y({\mathbf{x}}_i) > \frac{1+\rho_{max}}{2}$}
    {$\mathcal{S^*} \leftarrow \mathcal{S^*} \cup \{(\mathbf{x}^{distilled} = \mathbf{x}_i, \tilde{y} = \tilde{y}_i, \hat{y^*} = y)\}$}
    }

}
\textcolor{blue}{// Section.~\ref{sec:estimating}: Training Bayes Label Transition Network}

Minimize the $\hat{R}_1(\theta)$ in Eq.~\ref{eq:BayesNetwork} on $\mathcal{S^*}$ to learn the Bayes label transition network's parameter $\theta$.

\textcolor{blue}{// Section.~\ref{sec:training}: Training Classifier with Forward Correction}

Fix the learned $\theta$ and minimize the $\hat{R}_2(w)$ in Eq.~\ref{eq:clsTraining} on $\tilde{S}$ to learn the classifier's parameter $w$.

\KwOut{The classifier $f(\cdot;w)$}
\caption{Instance-dependent Label-noise Learning with Bayes Label Transition Network.}
\label{algo:learning}

\end{algorithm}

\subsection{Bayes Label Transition Network}\label{sec:estimating}

With the collected distilled examples $(\mathbf{x}^{distilled}, \tilde{y}, \hat{y^*})$, we proceed to train a Bayes label transition network parameterized by $\theta$ to estimate the instance-dependent Bayes label transition matrices $\hat{T_{i,j}^*}(\mathbf{x}^{distilled})$, which model the transition probabilities from Bayes optimal labels to noisy labels given input instances:
\begin{equation}
    \hat{T_{i,j}^*}(\mathbf{x}^{distilled};\theta)= P( \tilde{Y} = j | Y^* = i, \mathbf{x}^{distilled};\theta),
\end{equation}

where $\tilde{Y}$ indicates the noisy label and $Y^*$ indicates the Bayes optimal label. Specifically, the Bayes label transition network takes $\mathbf{x}^{distilled}$ as input and output an estimated Bayes-label transition matrix $\hat{T^*}(\mathbf{x}^{distilled};\theta) \in \mathbb{R}^{C \times C}$, where $C$ is the number of classes. 
We can use the collected Bayes labels $\hat{y^*}$ and the estimated Bayes-label transition matrix $\hat{T^*}(\mathbf{x}^{distilled}_i;\theta)$ to infer the noisy labels. The following empirical risk on the inferred noisy labels and the ground-truth noisy labels are minimized to learn the network's parameter $\theta$: 
\begin{equation}\label{eq:BayesNetwork}
    \hat{R}_1(\theta)=-\frac{1}{m}\sum_{i=1}^m \boldsymbol{\tilde{y}_i} \log (\boldsymbol{\hat{y^*_i}}\cdot \hat{T^*}(\mathbf{x}^{distilled}_i;\theta)),
\end{equation}
where $m$ is the number of distilled examples, $\boldsymbol{\tilde{y}_i}$ and $\boldsymbol{\hat{y^*_i}}$ are $\tilde{y}_i$ and $\hat{y^*_i}$ in the form of \textit{one-hot vectors}, $\boldsymbol{\tilde{y}_i} \in \mathbb{R}^{1 \times C}$ and $\boldsymbol{\hat{y^*_i}} \in \mathbb{R}^{1 \times C}$, respectively. Note that if we have a distilled example for the $i$-th class, we can only make use of it to learn the $i$-th row of the transition matrix. For the other rows, they will not contribute to calculate the loss of the current training example. However, it does not mean that they will be random or not learnable. Their information will be learned by exploiting distilled examples from the non-$i$-th classes. More specifically, the parameters of the network can be divided into row-specific parameters and commonly shared parameters. By assuming that we have distilled examples for each class, both the row-specific parameters and commonly shared parameters will be optimized.

\begin{table*}[!th]
	\centering
	
	\caption{Means and standard deviations (percentage) of classification accuracy on \textit{F-MNIST} with different label noise levels. `-V' indicates matrix revision~\citep{xia2019revision}. }
	\vspace{2mm}
	
	\label{tab:f-minist}
	{

		\resizebox{!}{0.24\linewidth}{\begin{tabular}{cccccc}
           	\toprule
			 ~& IDN-10\%      & IDN-20\%      & IDN-30\%    & IDN-40\%  & IDN-50\%         \\ \midrule
			 CE & 88.65 $\pm$ 0.45 & 88.31 $\pm$ 0.37 & 85.22 $\pm$ 0.56  & 76.56 $\pm$ 2.50 &  67.42 $\pm$ 3.91 \\
             GCE  & 90.86 $\pm$ 0.38 & 88.59 $\pm$ 0.26 & 86.64 $\pm$ 0.76 & 76.93 $\pm$ 1.64 & 66.69 $\pm$ 1.07  \\ 
		     APL & 86.46 $\pm$ 0.27 & 85.32 $\pm$ 0.88 & 85.59 $\pm$ 0.85 & 74.66 $\pm$ 2.77 & 62.82 $\pm$ 0.44\\
		     Decoupling & 89.83 $\pm$ 0.45 & 86.29 $\pm$ 1.13 & 86.01 $\pm$ 1.01 & 78.78 $\pm$ 0.53 & 67.33 $\pm$ 1.33\\
		     MentorNet &90.35 $\pm$ 0.64 & 87.92 $\pm$ 0.83 & 87.24 $\pm$ 0.99 & 79.01 $\pm$ 2.30 & 66.44 $\pm$ 2.97 \\
			 Co-teaching & 90.65 $\pm$ 0.58 & 88.77 $\pm$ 0.41 & 86.98 $\pm$ 0.67 & 78.92 $\pm$ 1.36 & 67.66 $\pm$ 2.42\\
			 Co-teaching+ & 90.47 $\pm$ 0.98 & 89.15 $\pm$ 1.77 & 86.15 $\pm$ 1.04 & 79.23 $\pm$ 1.30 & 63.49 $\pm$ 2.94\\
			 Joint & 80.19 $\pm$ 0.99 & 78.46 $\pm$ 1.24 & 72.73 $\pm$ 2.44 & 65.93 $\pm$ 2.08 & 50.93 $\pm$ 3.52\\
			 DMI & 91.58 $\pm$ 0.46 & 90.33 $\pm$ 0.66 & 85.96 $\pm$ 1.52& 77.77 $\pm$ 2.15 & 68.02 $\pm$ 1.59\\
			 Forward & 89.65 $\pm$ 0.24 & 88.61 $\pm$ 0.77 & 85.01 $\pm$ 0.43 & 78.59 $\pm$ 0.38 & 67.11 $\pm$ 1.46\\
			 Reweight & 90.33 $\pm$ 0.27 & 88.81 $\pm$ 0.44 & 84.93 $\pm$ 0.42 & 76.07 $\pm$ 1.93 & 67.66 $\pm$ 1.65\\
			 S2E & 91.04 $\pm$ 0.92 & 89.93 $\pm$ 1.08 & 86.77 $\pm$ 1.15 & 76.12 $\pm$ 1.21 & 70.24 $\pm$ 2.64 \\
			 T-Revision & 91.36 $\pm$ 0.59 & 90.24 $\pm$ 1.01 & 85.59 $\pm$ 1.77 & 78.24 $\pm$ 1.12 & 69.04 $\pm$ 2.92 \\
			 PTD & 92.03 $\pm$ 0.33 & 90.78 $\pm$ 0.64 & 87.86 $\pm$ 0.78 & 79.46 $\pm$ 1.58 & 73.38 $\pm$ 2.25\\
			 \midrule
			 BLTM & 96.06 $\pm$ 0.71 & 94.97 $\pm$ 0.33 &91.47 $\pm$ 1.36 & 82.88 $\pm$ 2.72 & 76.35 $\pm$ 3.79\\
			 BLTM-V & \textbf{96.93 $\pm$ 0.31} & \textbf{95.55 $\pm$ 0.59} &\textbf{92.24 $\pm$ 1.87} & \textbf{83.43 $\pm$ 1.72} & \textbf{76.89 $\pm$ 4.26}\\
             \bottomrule
		\end{tabular}}
	}
\end{table*}
\subsection{Classifier Training with Forward Correction}\label{sec:training}

Our goal is to train a classification network $f(\cdot|w)$ parameterized by $w$ that can predict Bayes class posterior probability $P(Y^* = i| \mathbf{x}; w)$. In the training stage, we cannot observe the Bayes optimal label $Y^*$. Instead, we only have access to noisy label $\tilde{Y}$. The probability of observing a noisy label $\tilde{Y}$ given input image $\mathbf{x}$ can be decomposed as:

\begin{align}
     \nonumber
     &P( \tilde{Y} = j \mid \mathbf{x};w,\theta) \\ 
    &= \sum_{i=1}^{k}{P( \tilde{Y} = j \mid Y^* = i, \mathbf{x};\theta)P( Y^* = i \mid \mathbf{x};w)},
\end{align}

With the trained Bayes label transition network, we can get $\hat{T^*_{i,j}}(\mathbf{x};\theta) = P( \tilde{Y} = j \mid Y^* = i, \mathbf{x};\theta)$ for each input $\mathbf{x}$. We exploit F-Correction \citep{patrini2017making}, which is a typical \textit{classifier-consistent} algorithm, to train the classification network. To be specific, fix the learned Bayes label transition network parameter $\theta$, we minimize the empirical risk as follows to optimize the classification network parameter $w$: 
\begin{equation}\label{eq:clsTraining}
    \hat{R}_2(w)=-\frac{1}{n}\sum_{i=1}^n \boldsymbol{\tilde{y}_i} \log (f(\mathbf{x}_i;w)\cdot \hat{T^*}(\mathbf{x}_i;\theta)),
\end{equation}

where $n$ is the number of all training examples in the noisy dataset and $f(\mathbf{x}_i;w) \in \mathbb{R}^{1 \times C}$. The F-Correction has been proved to be a classifier-consistent algorithm, the minimizer of $\hat{R}_2(w)$ under the noisy distribution is the same as the minimizer of the original cross-entropy loss under the Bayes optimal distribution~\citep{patrini2017making}, if the transition matrix $\hat{T^*}$ is estimated unbiased.
We show the overall framework in Algorithm.~\ref{algo:learning}, note the Bayes label transition network is trained on the distilled examples while the classifier is trained on the whole noisy dataset. The Bayes label transition network learned on the distilled examples will generalize to the non-distilled examples if they share the same pattern with the distilled examples which causes label noise. A recent study \citep{xia2020part} empirically verified that the patterns that cause label noise are commonly shared. Our empirical experiments further show that the network $\hat{T^*}(\mathbf{x};\theta)$ generalizes well to unseen examples and thus helps achieve superior classification performance.

\section{Experiments}\label{sec:4}
In this section, we first introduce the experiment setup (Section~\ref{sec:setup}) including the datasets used (Section~\ref{sec:datasets}), the implementation details (Section~\ref{sec:implement}), and the compared methods (Section~\ref{sec:baselines}). Then, we present and analyze the experimental results on synthetic and real-world noisy datasets to show the effectiveness of the proposed method (Section~\ref{sec:sota}). The noise generation algorithm, more comparison results, and more ablation studies are included in the Appendix.
\begin{table*}[!t]
	\centering
	\caption{Means and standard deviations (percentage) of classification accuracy on \textit{CIFAR-10} with different label noise levels. `-V' indicates matrix revision~\citep{xia2019revision}.}
	\vspace{2mm}
	\label{tab:cifar10}
	{
		\resizebox{!}{0.24\linewidth}{\begin{tabular}{cccccc}
           	\toprule
			 ~& IDN-10\%      & IDN-20\%      & IDN-30\%    & IDN-40\%  & IDN-50\%         \\ \midrule
			 CE &  73.54 $\pm$ 0.14 & 71.49 $\pm$ 1.35 & 67.52 $\pm$ 1.68 & 58.63 $\pm$ 4.92 & 51.54 $\pm$ 2.70\\
             GCE & 74.24 $\pm$ 0.89 & 72.11 $\pm$ 0.43 & 69.31 $\pm$ 0.18 & 56.86 $\pm$ 0.92 & 53.44 $\pm$ 1.28 \\ 
		     APL & 71.12 $\pm$ 0.19 & 68.89 $\pm$ 0.27 & 65.17 $\pm$ 0.35 & 53.22 $\pm$ 2.21 & 47.31 $\pm$ 1.41\\
		     Decoupling & 73.91 $\pm$ 0.37 & 74.23 $\pm$ 1.18 & 70.85 $\pm$ 1.88 & 54.73 $\pm$ 1.02 & 52.04 $\pm$ 2.09\\
		     MentorNet & 74.93 $\pm$ 1.37 & 73.59 $\pm$ 1.29 & 72.32 $\pm$ 1.04 & 57.85 $\pm$ 1.88 & 52.96 $\pm$ 1.98\\
			 Co-teaching & 75.49 $\pm$ 0.47 & 75.93 $\pm$ 0.87 & 74.86 $\pm$ 0.42 & 59.07 $\pm$ 1.03 & 55.62 $\pm$ 3.93\\
			 Co-teaching+ & 74.77 $\pm$ 0.16 & 75.14 $\pm$ 0.61 & 71.92 $\pm$ 2.13 & 59.15 $\pm$ 0.87 & 53.02 $\pm$ 3.34\\
			 Joint & 75.97 $\pm$ 0.98 & 76.45 $\pm$ 0.45 & 75.93 $\pm$ 1.65 & 63.22 $\pm$ 5.37 & 55.84 $\pm$ 3.25\\
			 DMI & 74.65 $\pm$ 0.13 & 73.49 $\pm$ 0.88 & 73.93 $\pm$ 0.34 & 60.22 $\pm$ 3.47 & 54.35 $\pm$ 2.28  \\
			 Forward & 72.35 $\pm$ 0.91 & 70.98 $\pm$ 0.32 & 66.53 $\pm$ 1.96 & 58.63 $\pm$ 1.25 & 52.33 $\pm$ 1.65\\
			 Reweight & 73.55 $\pm$ 0.32 & 71.49 $\pm$ 0.57 & 68.76 $\pm$ 0.37 & 60.32 $\pm$ 1.03 & 52.03 $\pm$ 1.70\\
			 S2E & 75.93 $\pm$ 1.01 & 75.53 $\pm$ 0.32 & 71.21 $\pm$ 2.51 & 64.62 $\pm$ 0.68 & 56.03 $\pm$ 1.07 \\
			 T-Revision & 74.01 $\pm$ 0.45 & 73.42 $\pm$ 0.64 & 71.15 $\pm$ 0.43 & 59.93 $\pm$ 1.33 & 55.67 $\pm$ 2.07\\
			 PTD & 76.33 $\pm$ 0.38 & 76.05 $\pm$ 1.72 & 75.42 $\pm$ 1.33 & 65.92 $\pm$ 2.33 & 56.63 $\pm$ 1.88\\
			 \midrule
			 BLTM  & 81.73 $\pm$ 0.56 & 80.26 $\pm$ 0.63 & 77.69 $\pm$ 1.37 & 71.96 $\pm$ 2.27 & 59.15 $\pm$ 3.11  \\
			 BLTM-V  & \textbf{82.16 $\pm$ 1.01} & \textbf{80.37 $\pm$ 1.98} & \textbf{78.82 $\pm$ 1.07} & \textbf{72.93 $\pm$ 4.00} & \textbf{60.33 $\pm$ 5.29}  \\
			 
             \bottomrule
		\end{tabular}}
	}
\end{table*}
\subsection{Experiment setup}
\label{sec:setup}
In this section, we introduce the four datasets we used to evaluate the proposed method, including three datasets with synthetic label-noise and one dataset with real-world label noise, and the baseline methods we compared with. 
\subsubsection{Datasets}
\label{sec:datasets}
We conduct the experiment on four datasets to verify the effectiveness of our proposed method, where three of them are manually corrupted, \textit{i.e.}, \textit{F-MNIST}, \textit{CIFAR-10}, and \textit{SVHN}, one of them is real-world noisy datasets, \textit{i.e.}, \textit{Clothing1M}. \textit{F-MNIST} has $28\times 28$ grayscale images of 10 classes including 60,000 training images and 10,000 test images. \textit{CIFAR-10} dataset contains 50,000 color images from 10 classes for training and 10,000 color images from 10 classes for testing both with shape of $32 \times 32 \times 3$. \textit{SVHN} has 10 classes of images with 73,257 training images and 26,032 test images. We manually corrupt the three datasets, \textit{i.e.}, \textit{F-MNIST}, \textit{CIFAR-10} and \textit{SVHN} with bounded instance-dependent label noise according to Algorithm~\ref{alg:noise_gen} (Appendix), which is modified from~\citep{xia2020part}. In noise generation, the noise rate upper bound $\rho_{max}$ in Algorithm~\ref{alg:noise_gen} is set as 0.6 for all experiments. All experiments on those datasets with synthetic instance-dependent label noise are repeated five times to guarantee reliability. The \textit{Clothing1M} has 1M images with real-world noisy labels for training and 10k images with the clean label for testing, only noisy samples are exploited to train and validate the model. 10\% of the noisy training examples of all datasets are left out as a noisy validation set for model selection. 

\subsubsection{Implementation details}
\label{sec:implement}
We use ResNet-18~\citep{he2016deep} for \textit{F-MNIST}, ResNet-34 networks~\citep{he2016deep} for \textit{CIFAR-10} and \textit{SVHN}. We first use SGD with momentum 0.9, batch size 128, and an initial learning rate of 0.01 to warm up the network for five epochs on the noisy dataset. For \textit{Clothing1M}, we use a ResNet50 pretrained on ImageNet, and the learning rate is set as 1e-3. Then, we use the warm-upped network to collect distilled examples from noisy datasets according to Section~\ref{sec:collecting}. The noise rate upper bound $\rho_{max}$ in Algorithm.~\ref{algo:learning} is manually set to 0.3 for all experiments to avoid laborious tuning, we show the distillation quality with difference choice of $\rho_{max}$ in the Appendix.  After distilled examples collection, we train the Bayes label transition network on the distilled dataset for 5 epochs. For model design brevity, we keep the architecture of the Bayes label transition network as the same as the architecture of the classification network, but the last linear layer is modified according to the transition matrix shape.  The optimizer of the Bayes label transition network is SGD, with a momentum of 0.9 and a learning rate of 0.01. Then, we fix the trained Bayes label transition network to train the classification network. The Bayes label transition network is used to generate a transition matrix for each input image; the transition matrix is used to correct the outputs of the classification network to bridge the Bayes posterior and the noisy posterior. The classification network is trained on the noisy dataset for 50 epochs for \textit{F-MNIST}, \textit{CIFAR-10} and \textit{SVHN} and for 10 epochs for \textit{Clothing1M} using Adam optimizer with a learning rate of $5e-7$ and weight decay of $1e-4$. We also apply the transition matrix revision technique~\citep{xia2019revision} to boost the performance. Note for a fair comparison, we do not use any data augmentation technique in all experiments as in~\citep{xia2020part}. All the codes are implemented in PyTorch 1.6.0 with CUDA 10.0, and run on NVIDIA Tesla V100 GPUs.
\begin{table*}[!t]
	\centering
	\caption{Means and standard deviations (percentage) of classification accuracy on \textit{SVHN} with different label noise levels. `-V' indicates matrix revision~\citep{xia2019revision}. }
	\vspace{2mm}
	\label{tab:svhn}
	{
		\resizebox{!}{0.24\linewidth}{\begin{tabular}{cccccc}
           	\toprule
			 ~& IDN-10\%      & IDN-20\%      & IDN-30\%    & IDN-40\%  & IDN-50\%         \\ \midrule
			 CE & 90.39 $\pm$ 0.13 & 89.04 $\pm$ 1.32 & 85.65 $\pm$ 1.84 & 79.94 $\pm$ 2.71 & 61.01 $\pm$ 5.41 \\
             GCE  & 90.82 $\pm$ 0.15 & 89.35 $\pm$ 0.94 & 86.43 $\pm$ 0.63 & 81.66 $\pm$ 1.58 & 54.77 $\pm$ 0.25  \\ 
		     APL & 71.78 $\pm$ 0.76 & 89.48 $\pm$ 1.67 & 83.46 $\pm$ 2.17 & 77.90 $\pm$ 2.31 & 55.25 $\pm$ 3.77\\
		     Decoupling & 90.55 $\pm$ 0.83 & 88.74 $\pm$ 0.77 & 85.03 $\pm$ 1.63 & 83.36 $\pm$ 2.73 & 56.76 $\pm$ 1.87\\
		     MentorNet & 90.28 $\pm$ 0.52 & 89.09 $\pm$ 0.95 & 85.89 $\pm$ 0.73 & 82.63 $\pm$ 1.73 & 55.27 $\pm$ 4.14 \\
			 Co-teaching & 91.05 $\pm$ 0.33 & 89.56 $\pm$ 1.77 & 87.75 $\pm$ 1.37 & 84.92 $\pm$ 1.59 & 59.56 $\pm$ 2.34\\
			 Co-teaching+ & 92.83 $\pm$ 0.87 & 90.73 $\pm$ 1.39 & 86.37 $\pm$ 1.66 & 75.24 $\pm$ 3.77 & 54.58 $\pm$ 3.46\\
			 Joint & 88.39 $\pm$ 0.62 & 85.37 $\pm$ 0.44 & 81.56 $\pm$ 0.43 & 78.98 $\pm$ 2.98 & 59.14 $\pm$ 3.22 \\
			 DMI & 92.11 $\pm$ 0.49 & 91.63 $\pm$ 0.87 & 86.98 $\pm$ 0.36 & 81.11 $\pm$ 0.68 & 63.22 $\pm$ 3.97\\
			 Forward & 90.01 $\pm$ 0.78 & 89.77 $\pm$ 1.54 & 86.70 $\pm$ 1.44 & 80.24 $\pm$ 2.77 & 57.57 $\pm$ 1.45\\
			 Reweight & 91.06 $\pm$ 0.19 & 92.01 $\pm$ 1.04 & 87.55 $\pm$ 1.71 & 83.79 $\pm$ 1.11 & 55.08 $\pm$ 1.25\\
			 S2E &  92.70 $\pm$ 0.51 & 92.02 $\pm$ 1.54 & 88.77 $\pm$ 1.77 & 83.06 $\pm$ 2.19 & 65.39 $\pm$ 2.77\\
			 T-Revision & 93.07 $\pm$ 0.79 & 92.67 $\pm$ 0.88 & 88.49 $\pm$ 1.44 & 82.43 $\pm$ 1.77 & 67.64 $\pm$ 2.57\\
			 PTD & 93.77 $\pm$ 0.33 & 92.59 $\pm$ 1.07 & 89.64 $\pm$ 1.98 & 83.56 $\pm$ 2.21 & 71.57 $\pm$ 3.32\\
			 \midrule
			 BLTM & 96.05 $\pm$ 0.32 & 94.97 $\pm$ 0.58 & 93.99 $\pm$ 1.24 & 87.67 $\pm$ 1.29 & 78.13 $\pm$ 4.62\\
			 BLTM-V & \textbf{96.37 $\pm$ 0.77} & \textbf{95.12 $\pm$ 0.40} & \textbf{94.69 $\pm$ 0.24} & \textbf{88.13 $\pm$ 3.23} & \textbf{78.71 $\pm$ 4.37}\\
             \bottomrule
		\end{tabular}}
	}
\end{table*}
\begin{table*}[!h]

	\centering
	\caption{Classification accuracy on \textit{Clothing1M}. In the experiments, only noisy samples are exploited to train and validate the deep model.}
\vspace{1mm}
	{
		\resizebox{!}{0.075\linewidth}{\begin{tabular}{cccccccc}
           	\toprule
			 CE &Decoupling & MentorNet & Co-teaching & Co-teaching+ & Joint & DMI \\ \midrule
			 68.88 & 54.53 & 56.79 & 60.15 & 65.15 & 70.88 & 70.12  \\ \toprule
			     Forward & Reweight & T-Revision & PTD &PTD-V  & BLTM & BLTM-V\\ \midrule
			    69.91 & 70.40 & 70.97 & 70.07 & 70.26  & \textbf{73.33} & \textbf{73.39}	\\ \toprule   
		\end{tabular}}
	}
	\label{tab:clothing1m}
\end{table*}
\subsubsection{Comparison Methods}
\label{sec:baselines}
We compare the proposed method with several state-of-the-art approaches: (1) CE, which trains the classification network with the standard cross-entropy loss on noise datasets. (2) GCE~\citep{zhang2018generalized}, which unites the mean absolute error loss and the cross-entropy loss to combat noisy labels. (3) APL~\citep{APL}, which combines two mutually reinforcing robust loss functions, we employ its combination of NCE and RCE for comparison. (4) Decoupling~\citep{malach2017decoupling}, which trains two networks on samples whose predictions from two networks are different. (5) MentorNet~\citep{jiang2018mentornet}, Co-teaching~\citep{han2018co}, and Co-teaching+~\citep{yu2019does} mainly handle noisy labels by training networks on instances with small loss values. (6) Joint~\citep{tanaka2018joint}, which jointly optimizes the network parameters and the sample labels. The hyperparameters $\alpha$ and $\beta$ are set to 1.2 and 0.8, respectively. (7) DMI~\citep{xu2019l_dmi}, which proposes a novel information-theoretic loss function for training neural networks robust to label noise. (8) Forward~\citep{patrini2017making}, Reweight~\citep{liu2016classification}, and T-Revision~\citep{xia2019revision} utilize a class-dependent transition matrix $T$ to correct the loss function. (9) PTD~\citep{xia2020part}, estimates instance-dependent transition matrix by combing part-dependent transition matrices, which is the most related work to our proposed method. We also provide comparison results between our method and DivideMix\citep{li2020dividemix} in Appendix, which is a hybrid algorithm that combines multiple powerful techniques, e.g. Gaussian Mixture Model, MixMatch, MixUp, regularization and asymmetric noise penalty in Appendix. As for our method, we simply model the instance-dependent matrix by employing a neural network.

\subsection{Comparison with the State-of-the-Arts}
\label{sec:sota}

\noindent\textbf{Results on synthetic noisy datasets.} Table~\ref{tab:f-minist},\ref{tab:cifar10} and \ref{tab:svhn} report the classification accuracy on the datasets of \textit{F-MNIST},\textit{CIFAR-10}, and \textit{SVHN}, respectively.

For \textit{F-MNIST}, our method surpasses all the baseline methods by a large margin. Equipping the transition matrix revision (-V)~\citep{xia2019revision} can further boost the performance of our method. For \textit{SVHN} and \textit{CIFAR-10}, the superiority of our method is gradually revealed along with the noise rate increase, which shows that our method can handle the extremely hard situation much better. Specifically, the classification accuracy of our method is 5.83\% higher than PTD (the best statistically consistent baseline) on \textit{CIFAR-10} in the IDN-10\% case, and the performance gap is enlarged to 7.01\% in the IDN-40\% case. On the \textit{SVHN}, the classification accuracy of our method is 2.60\% higher than PTD in the IDN-10\% case, 5.05\% higher than PTD in the IDN-30\% case, and 7.14\% higher than PTD in the most challenging IDN-50\% case. 
=

\noindent\textbf{Results on real-world noisy datasets.}
The noise model of real-world datasets is more likely to be instance-dependent. By extracting Bayes optimal labels and explicitly learning the noise transition patterns on the challenging \textit{Clothing1M}, a dataset with real human label noise, our proposed method also performs favorably well, which proves that our method is more flexible to handle such real-world noise problem.

\section{Conclusion}\label{sec:conclusion}
In this paper, we focus on training the robust classifier with the challenging instance-dependent label noise. To address the issues of existing \textit{clean-label transition matrix}, we propose to directly build the transition between \textit{Bayes optimal labels} and \textit{noisy labels}. By reducing the feasible solution space of the transition matrix estimation, we prove that the instance-dependent label transition matrix that relates \textit{Bayes optimal labels} and \textit{noisy labels} can be directly learned using \textit{deep neural networks}. Experimental results demonstrate that the proposed method is more superior in dealing with instance-dependent label noise, especially for the case of high-level noise rates.

\section{Acknowledgements}
TLL is partially supported by Australian Research Council Projects DE-190101473, IC-190100031, and DP-220102121. EKY is partially supported by Guangdong Basic and Applied Basic Research Foundation (2021A1515110026) and Natural Science Basic Research Program of Shaanxi (Program No.2022JQ-608). YL is partially supported by the National Science Foundation (NSF) under grants IIS-2007951, IIS-2143895, and CCF-2023495. BH is partially supported by the RGC Early Career Scheme No. 22200720, NSFC Young Scientists Fund No. 62006202, and Guangdong Basic and Applied Basic Research Foundation No. 2022A1515011652.

\bibliography{ref}
\bibliographystyle{icml2022}
\clearpage

\appendix
\onecolumn

\begin{algorithm*}[t!]
\caption{Bounded Instance-dependent Label Noise Generation.}
\label{alg:noise_gen}
{
\textbf{Required}: Clean examples $\{({\mathbf{x}}_i, y_i)\}_{i=1}^{n}$; Noise rate $\eta$; Noise rate upper bound $\rho_{max}$;

Sample instance flip rates $q_i$ from the truncated normal distribution $\mathcal{N}(\eta,0.1^2,[0,\rho_{max}])$;\\
\textcolor{blue}{\hfill//mean $\eta$, variance $0.1^2$, range [0,$\rho_{max}$]}

Independently sample $w_1,w_2,\ldots,w_c$ from the standard normal distribution $\mathcal{N}(0,1^2)$;

\For{$i=1,2,\ldots,n$} 
{
$p={\mathbf{x}}_i\times w_{y_i}$;\textcolor{blue}{\hfill//generate instance-dependent flip rates}

$p_{y_i}=-\infty$;\textcolor{blue}{\hfill//only consider entries that are different from the true label}

$p=q_i\times softmax(p)$;\textcolor{blue}{\hfill//make the sum of the off-diagonal entries of the $y_i$-th row to be $q_i$}

$p_{y_i}=1-q_i$;\textcolor{blue}{\hfill//set the diagonal entry to be 1-$q_i$}

Randomly choose a label from the label space according to the possibilities $p$ as noisy label $\tilde{y}_i$;
}

\KwOut{Noisy samples $\{({\mathbf{x}}_i, \tilde{y}_i)\}_{i=1}^{n}$
}}

\end{algorithm*}

\section{Hyper-parameter Sensitivity}
\label{sec:sense}
The quality of the distilled dataset relies on the choice of distillation threshold $\hat{\rho}_{max}$ (denoted as $\hat{\rho}$ in the following paragraph) in Algorithm~\ref{algo:learning}. The distillation threshold $\hat{\rho}$ controls how many examples can be collected out of noisy dataset and the distillation accuracy. If the $\hat{\rho}$ is not smaller than the ground-truth noise rate, all collected Bayes optimal labels are theoretically guaranteed~\cite{cheng2017learning}. We analyse the effect of $\hat{\rho}$ on the CIFAR-10 dataset in Table.~\ref{tab:quality}. The distillation accuracy is computed by counting how many inferred Bayes optimal labels are consistent with their corresponding true labels among all distilled examples. 

\begin{figure*}
    \centering
    \resizebox{!}{0.5\linewidth}{\includegraphics{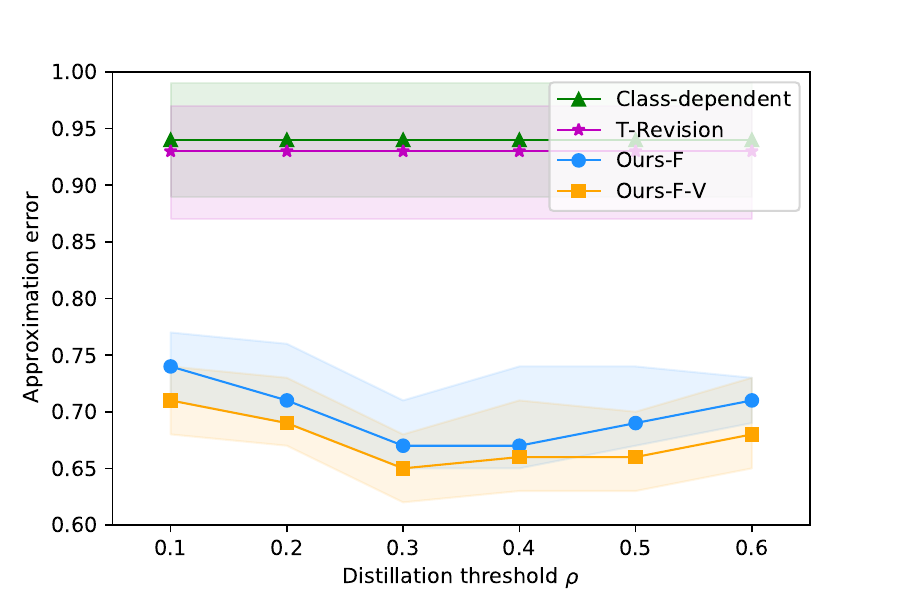}}
    \caption{Illustration of the transition matrix approximation error on CIFAR-10 with IDN-30\% noise rate. The error bar for standard deviation has been shaded.} 
    \label{fig:appro}
\end{figure*}

In Figure~\ref{fig:appro}, we show the instance-dependent transition matrix approximation error when employing the class-dependent transition matrix, the revised class-dependent transition matrix, and our proposed instance-dependent transition matrix estimation method. The error is measured by $\ell_1$ norm between the ground-truth transition matrix and the estimated transition matrix. For each instance, we only analyze the approximation error of a specific row because the noisy label is generated by one row of the instance-dependent transition matrix. The "Class-dependent" represents the class-dependent transition matrix learning methods~\citep{patrini2017making}, the `T-Revision' indicates the class-dependent transition matrix is revised by a learnable slack variable~\citep{xia2019revision}. Our proposed method estimates an instance-dependent transition matrix for each input. It can be observed that our proposed method can achieve a much lower approximation error. 

We manually set $\hat{\rho} = 0.3$, a decent trade-off between distillation accuracy and the number of distilled examples, in all experiments to avoid laborious hyper-parameter tuning and accessing to the true noise rate.

\begin{table}[tph]
  \centering
{
  \begin{tabular}{lcccc}
  \hline
    \multirow{2}{*}{Noise rate} & \multicolumn{2}{c}{$\hat{\rho} = 0.3$} & \multicolumn{2}{c}{$\hat{\rho} = 0.5$} \\
     \cline{2-3} \cline{4-5} 

        & distill. acc. & \# of distilled examples & distill. acc. & \# of distilled examples \\
    \hline
    IDN-10\%  & 98\% & 27983 / 50000 & 99\% & 19983  / 50000\\
    IDN-30\%  & 96\% & 17673  / 50000 & 99\% & 10673  / 50000 \\
    IDN-50\%. & 94\% & 8029   / 50000 & 98\% & 5098  / 50000 \\
    \hline
  \end{tabular}}
  \caption{Distillation quality analysis on CIFAR-10, with total 50,000 examples in the original non-distilled dataset.   } \label{tab:quality}
\end{table}

\section{Ablation on Bayes-label transition matrix}

To verify the effectiveness of the estimated Bayes-label transition matrix, we compare our method with some ablated variants, e.g. directly train a classifier on the distilled dataset and relabel the noisy dataset using the classifier trained on distilled dataset. 

\begin{table*}[h!]
	\centering
	{
		\begin{tabular}{lccccc}
          	\toprule
			 ~& CIFAR-10 IDN-10\%      & CIFAR-10 IDN-30\%  & Clothing1M        \\ \midrule
			 Training classifier on distilled dataset & 74.56 & 67.42 & 62.37 \\
			 Relabeling noisy dataset & 76.68 & 70.73 & 64.98 \\ 
			 Ours  & \textbf{82.16} & \textbf{78.82} & \textbf{73.39}\\
             \bottomrule
		\end{tabular}
	}
\end{table*}

\section{Comparision with DivideMix}
DivideMix has a much more complicated pipeline than us and is not a statistically consistent algorithm. We compare our method with DivideMix to further show the effectiveness and flexibility of our proposed method. Compared with DivideMix, our method exhibit competitive performance when noise rate is low and surpass DivideMix by a large margin on the worst noise cases (3.22\% performance improvement on CIFAR-10 and 4.38\% on SVHN, both under IDN-50\% ), with a much simpler and flexible algorithm design.
\begin{table*}[pht!]
	\centering
	{
		\begin{tabular}{lccccc}
          	\toprule
			 ~& IDN-10\%  &  IDN-20\% &  IDN-30\% &  IDN-40\% &  IDN-50\%    \\ \midrule
			 DivideMix & 96.37 $\pm$ 0.72 & 95.92 $\pm$ 0.73 & 90.37 $\pm$ 0.83 & 80.92 $\pm$ 2.32 & 74.63 $\pm$ 3.76 \\
			 Ours & \textbf{96.93 $\pm$ 0.31} & \textbf{95.55 $\pm$ 0.59} &\textbf{92.24 $\pm$ 1.87} & \textbf{83.43 $\pm$ 1.72} & \textbf{76.89 $\pm$ 4.26}\\
             \bottomrule
		\end{tabular}
	}
	\caption{F-MNIST}
	
\end{table*}
\begin{table*}[pht!]
	\centering
	{
		\begin{tabular}{lccccc}
          	\toprule
          	~& IDN-10\%  &  IDN-20\% &  IDN-30\% &  IDN-40\% &  IDN-50\%    \\ \midrule
			  DivideMix & \textbf{83.31 $\pm$ 0.23} & \textbf{81.42 $\pm$ 0.28} & \textbf{80.73 $\pm$ 1.28} & 70.29 $\pm$ 1.97 & 57.11 $\pm$ 3.64 \\ 

			 Ours  & {82.16 $\pm$ 1.01} & {80.37 $\pm$ 1.98} & {78.82 $\pm$ 1.07} & \textbf{72.93 $\pm$ 4.00} & \textbf{60.33 $\pm$ 5.29}  \\
             \bottomrule
		\end{tabular}
	}
	\caption{CIFAR10}
	
\end{table*}
\begin{table*}[pht!]
	\centering
	{
		\begin{tabular}{lccccc}
          	\toprule
          	~& IDN-10\%  &  IDN-20\% &  IDN-30\% &  IDN-40\% &  IDN-50\%    \\ \midrule
			  DivideMix & 96.02 $\pm$ 0.45 & \textbf{95.73 $\pm$ 0.48} & 92.07 $\pm$ 1.47 & 85.69 $\pm$ 2.47 & 74.33 $\pm$ 4.07 \\
			 Ours & \textbf{96.37 $\pm$ 0.77} & {95.12 $\pm$ 0.40} & \textbf{94.69 $\pm$ 0.24} & \textbf{88.13 $\pm$ 3.23} & \textbf{78.71 $\pm$ 4.37}\\
             \bottomrule
		\end{tabular}
	}
	\caption{SVHN}
	
\end{table*}

\end{document}